\pdfoutput=1

\documentclass[11pt]{article}

\usepackage[preprint]{acl}

\usepackage{tabularx}
\usepackage{booktabs}
\usepackage{longtable}
\usepackage{tabularray}
\usepackage{times}
\usepackage{latexsym}
\usepackage{CJKutf8}
\usepackage{tabularx}
\usepackage{booktabs}
\usepackage[T1]{fontenc}

\usepackage[utf8]{inputenc}

\usepackage{microtype}

\usepackage{inconsolata}
\usepackage{graphics}
\usepackage{graphicx}
\usepackage{hyperref} 
\usepackage{tabularray}
\usepackage{multirow}
\usepackage{arydshln}
\usepackage{multirow}
\usepackage{colortbl}
\usepackage{color}
\usepackage{tabularray}
\usepackage{array}
\usepackage{multirow}
\usepackage{ragged2e}
\usepackage{colortbl}
\usepackage{hhline}

\title{Probing Causality Manipulation of Large Language Models}

\author{Chenyang Zhang~\thanks{Equally Contribution.}, Haibo Tong~\footnotemark[1], Bin Zhang, Dongyu Zhang \\
\\
  Tongji University \\
  \\
  \texttt{\{inkzhangcy,2151130,2233009,yidu\}@tongji.edu.cn} \\
  \\}
\begin{document}

\maketitle
\begin{abstract}
  Large language models (LLMs) have shown various ability on natural language processing, including problems about causality. It is not intuitive for LLMs to command causality, since pretrained models usually work on statistical associations, and do not focus on causes and effects in sentences. So that probing internal manipulation of causality is necessary for LLMs. This paper proposes a novel approach to probe causality manipulation hierarchically, by providing different shortcuts to models and observe behaviors. We exploit retrieval augmented generation (RAG) and in-context learning (ICL) for models on a designed causality classification task. We conduct experiments on mainstream LLMs, including GPT-4 and some smaller and domain-specific models. Our results suggest that LLMs can detect entities related to causality and recognize direct causal relationships. However, LLMs lack specialized cognition for causality, merely treating them as part of the global semantic of the sentence.~\footnote{Our code and implementation are available at \href{https://github.com/TongjiNLP/llm-causality-probing}{https://github.com/TongjiNLP/llm-causality-probing}.}
\end{abstract}

\section{Introduction}

Large language models (LLMs) exhibit a diverse range of capabilities in Natural Language Processing (NLP)~\citep{wei2022emergent,DBLP:conf/fat/GanguliHLABCCDD22}.
Though LLMs are still based on statistical machine learning~\citep{DBLP:books/acm/22/BareinboimCII22,chen-etal-2023-say}, they behave well in some inference and reasoning tasks~\citep{DBLP:conf/iclr/BhagavatulaBMSH20}, showing ability for manipulation of \textbf{causality}.

However, intrinsic manipulation of causality remains unclear for researchers.
Unfortunately, investigating intrinsic manipulation is not straightforward for LLMs due to complex model structure.
They have enormous parameters, magnifying the cost of refactoring models.
And more advanced architectures like Mixture-of-Experts (MoE)~\citep{deepseekai2024deepseek,jiang2023mistral} proposes challenge for detailed probing, because behaviors of models are hard to guide.
Moreover, some existing models do not share technical details.
Intuitive research like ablation study is hard to work under such circumstances.

\begin{figure*}[!ht]
	\begin{center}
		\includegraphics[width=1.0\linewidth]{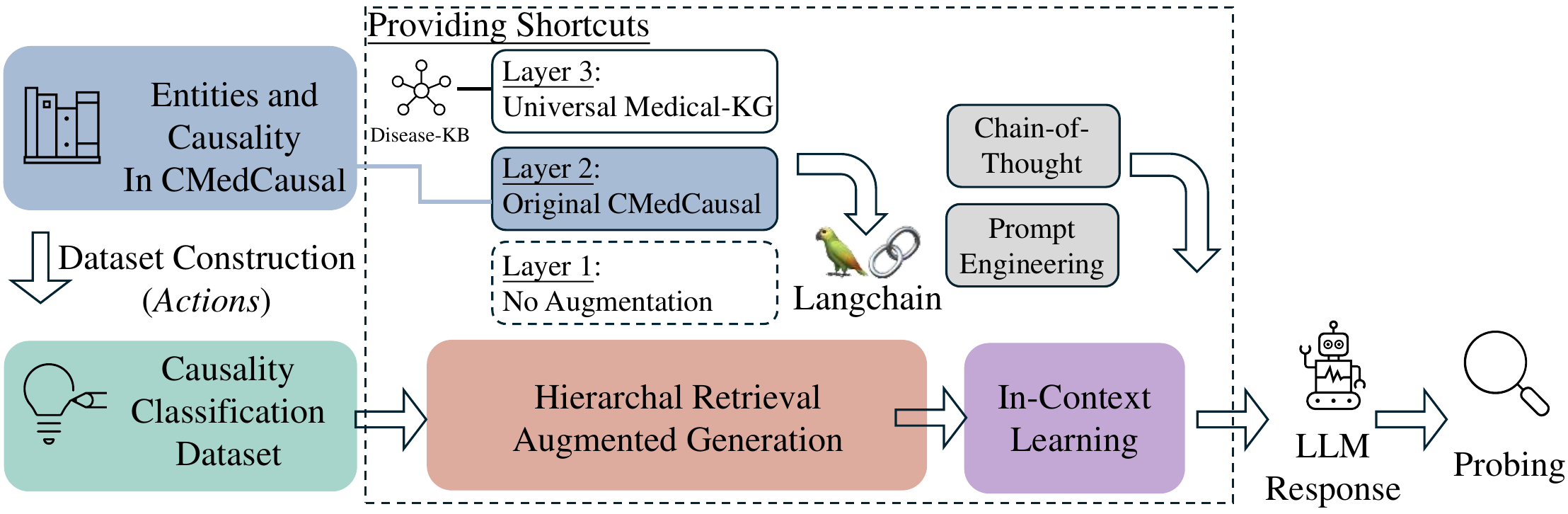}
		\caption{Main stricture of our probing works. We construct a causal dataset, then guide models by providing shortcuts. Finally, we probe intrinsic manipulation of causality by comparing performances of different shortcuts.}
		\label{fig:IntroFig}
	\end{center}
\end{figure*}

To address this challenge, our work proposes an innovative approach of probing intrinsic manipulation of causality for LLMs.
As shown in Fig.~\ref{fig:IntroFig}, firstly we construct a classification dataset for detecting entities and relationships of causality in sentences.
Then we guide behaviors of LLMs by hierarchically add \textit{shortcuts} on this classification task.
We integrate retrieval augmented generation (RAG) and in context learning (ICL) for providing shortcuts.
This takes into account the effects of prompts and pretrained knowledge into consideration while probing.
Finally, we observe performance variance under different RAG and ICL, to probe intrinsic manipulation of causality.
We conduct experiments on LLMs in various parameters sizes and domain knowledge.
The experimental results show that LLMs are sensitive to global semantics in classification, and show a certain ability to identify causal entities with guidance. But they do not have direct cognition of causal relationships, lacking a fixed processing route for causality. This leads to sub-optimal performance in more complex problem scenarios for causality, indicating necessity  for further attention in LLMs' training.

\section{Related Work}
\subsection{Probing LLMs}
The working mechanisms of LLMs remain unclear, raising concerns about the reliability and effectiveness of their generated content.
Probing~\citep{hewitt-manning-2019-structural} aims to discern the internal behaviors of models.
Probing researches on LLMs have offered valuable insights into various topics,
like mathematical~\citep{stolfo-etal-2023-causal}, sociology~\citep{ramezani-xu-2023-knowledge,hossain-etal-2023-misgendered}, and pretrained knowledge~\citep{chen-etal-2023-say}.
In context learning~\citep{brown2020language} is common approach in probing LLMs, since it enables guidance of LLMs without additional training.
Furthermore, furnishing models with specific knowledge has been proven to be an effective probing strategy~\citep{lin-etal-2020-commongen,chen-etal-2023-say}.
\subsection{Evaluation of Causality for LLMs}
Causality in LLMs has been explored through tasks like commonsense inference~\citep{DBLP:conf/iclr/BhagavatulaBMSH20,talmor-etal-2019-commonsenseqa}, event causal identification~\citep{gao-etal-2019-modeling,mu-li-2023-enhancing}, and explanation generation~\citep{du-etal-2022-e}, with ChatGPT's abilities evaluated~\citep{gao-etal-2023-chatgpt}. 

However, the integration of causality in real-world domains~\citep{CausalFrontier} contrasts with LLMs' reliance on statistical associations~\citep{CausalParrots}. Furthermore,~\citep{jin2024can} confirms LLMs' lack of causal reasoning, pointing towards a gap in theoretical discussion despite practical applications.

\section{Dataset Construction}\label{sec:dataset}
In this section, we introduce an innovative approach to construct a classification dataset for probing.
Our approach focuses on entities and their causal relationships in sentence, and diminishes interference of pretrained knowledge for probing.
Moreover, our approach preserve gold standard for the classification tasks, which is feasible for providing "shortcuts" and to guide behaviours of models.
\begin{figure}[!ht]
	\begin{center}
		\includegraphics[width=1.0\columnwidth]{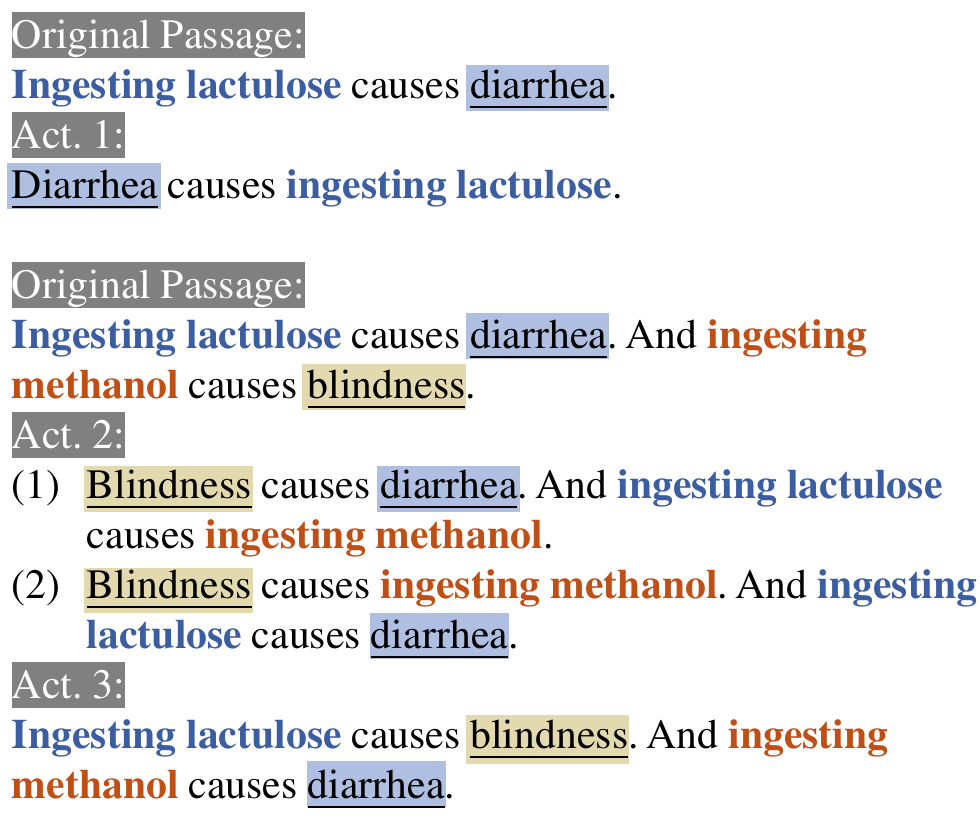}
		\caption{An instance of our constructed datasets.  Causes in sentences are bold and effects are underlined. The corresponding causes and effects are marked with the same color.}
		\label{fig:Sec3_dataset_illustration}
	\end{center}
\end{figure}
\paragraph{Base Dataset}
We construct our dataset based on the CMedCausal dataset~\citep{zhang-etal-2022-cblue}.
CMedCausal provides medical expressions and annotates all causal relationships and  entities (causes and effects) in sentences.
More details and cases about base dataset can be found in Appendix~\ref{sec:Apd_dataset_details}.

\paragraph{Classification Dataset Construction}
For classification, we sample original sentences from CMedCausal as \textbf{positive instances}, as they contain correct causation.
And we produce \textbf{negative instances} with certain manners (notated as \textit{Actions} or \textit{Act.}).
Trough different actions, causation between entities are disturbed, but other parts of sentences are preserved in the best effort.
Fig.~\ref{fig:Sec3_dataset_illustration} gives instance of three actions.

\textbf{Action 1: Local Causation Disturbing}
We swap the order of cause and effect~\footnote{Cause and effect of CMedCausal usally contains medical named entities.} in positive instance, to probe model manipulation of single causation in sentences.
We filter the corresponding original texts with a limited length.
Passages are segmented using a Chinese full stop character, and we select minimum continuous sentence sequence~\footnote{The sequence starts with first sentence contains causal mentions and ends with the last sentence contains causal mentions.} containing modified parts.

\textbf{Action 2: Global Causation Disturbing}
This action introduces a stronger disturbance for global semantics.
We shuffle all entities mentioned in any causation (no matter they are causes or effects), and put entities in shuffled order to produce a negative instance.

\textbf{Action 3: Mutual Causation Disturbing}
This action delves into the model's further understanding of causation, specifically focusing on interactions between causation, which is based on cognition of causation.
We select two sentences with causation, pinpoint one causation in each sentence and swap another. For example, $A\to B$ (represents \textit{$A$ causes $B$}) and $C \to D$ are swapped to yield $A\to D$ and $C\to B$.
And then altered causation is placed into original sentences to produce negative instances.

\section{Probing Design}
In this section, we propose a hierarchical probing approach.
As shown in Fig.~\ref{fig:IntroFig}, our method provides "shortcuts" hierarchically to LLMs in classification tasks.
These shortcuts include necessary steps for causality manipulation, like entities recognition and alignment, causal relation cognition.
By comparing whether these shortcuts are beneficial to tasks performance, intrinsic manipulation of causality is probed.
We exploit a combination of RAG and ICL for providing shortcuts to guide LLMs.
For evaluation, we rewrite classification tasks in Sec.~\ref{sec:dataset} into a question and answer form, requesting LLMs to judge whether causality of the sentence is right.

\subsection{Hierarchical Retrieval Augmented Generation}
We add different augmentation for LLMs, forming a hierarchical structure in Fig.~\ref{fig:IntroFig}, notated as \textit{layers}.
For each layer, we retrieve most relevant sentences and attach them in questions for LLMs.
From layer 1 to layer 3, we provide more complex guidance from shortcuts, representing a more ideal and detailed manipulation of causality. And we aim to probe whether models show identical manipulation as guided, which can be observed from performance changes.

\paragraph{Layer 1: No Augmentation} This layer offers no augmentation for LLMs, to demonstrate models native manipulation.

\paragraph{Layer 2: Original CMedCausal}

This layer provides the most efficient shortcuts, that is, the original passages used in dataset construction.
These shortcuts are derived from the original CMedCausal, serving as gold standard for classification. Consequently, this layer guides models to infer about basic causality, probing causal entities recognition and causality understanding.

Additionally, we exploit back-translation for this layer, notated as \textbf{Layer 2.5: Original CMedCausal (back-translated)}, implementation details can be found in Appendix.~\ref{sec:BackTranslation}.

\paragraph{Layer 3: Universal Medical-KG}

This classification dataset is in medical domain w common diseases in Chinese.
And we supplement the necessary medical knowledge, aiming to guide models to infer latent causality in sentences. LLMs are required to recognize entities and derive causality in knowledge. 
To provide proper medical knowledge, we use a Chinese common disease knowledge graph, DiseaseKG\footnote{\href{https://github.com/nuolade/disease-kb}{https://github.com/nuolade/disease-kb}}. We discuss about effectiveness of augmented knowledge in Appendix.~\ref{sec:Effectiveness_Layer3}.

\paragraph{Retrieval Augmented Generation}
To extract medical information from a large corpus, we adopt a retriever-reader pipeline~\citep{chen-etal-2017-reading}. By integrating the retrieved knowledge with the questions, the model can gain more medical expertise, enhancing the accuracy of its answers. Additionally, efforts should be made to minimize the influence of specialized knowledge on the model's ability to discern causality. The specific method of retrieval can be referred to in Appendix~\ref{sec:langchain}.
\subsection{In Context Learning Design}
In this section, we mainly exploit ICL for guidance of LLMs.
In detail, our main approach include prompt engineering.
Moreover, we integrate chain of thought~\citep{DBLP:conf/nips/KojimaGRMI22,DBLP:conf/iclr/ZhouSHWS0SCBLC23,DBLP:conf/nips/Wei0SBIXCLZ22} in prompts as further shortcuts.

We provide prompts with necessary information, following a prompt framework in community \footnote{\href{https://www.promptingguide.ai/}{https://www.promptingguide.ai/}}.
We do not conduct further prompt engineering, since we believe that LLMs should comprehends natural prompts.
Prompt includes instructions, contexts, input data and output indicator, introduction of these components can be found in Appendix.~\ref{Structure_of_Prompts}.

The \textbf{simple prompt} provides components mentioned above, but no additional guidance. It is used to probe native thinking process directly.
In order to better exploit causality ability of LLMs, we use \textbf{advanced prompt} for additional experiments, indicating an upper bound.
Advanced prompt integrates chain of thoughts similar to \citep{DBLP:conf/nips/Wei0SBIXCLZ22}, which  prompts models with types of mistakes (e.g. wrong orders of entities in causality), and instruct models to give an explanation and then conduct classification.
Since final sentences concatenated will be very long, we append some part of sentence (i.e. knowledge provided) into history with a multiple rounds dialog.
Examples of simple prompt and advanced prompt can be found in Appendix~\ref{Example_of_Prompts}.

\section{Experiments}
\subsection{Experiment Settings}
\paragraph{Model Selection}
During models selection, language preference, domains preference, parameters size and feasibility of probing are considered during models selection.
So we select following models:

\textbf{GPT-4}~\citep{DBLP:journals/corr/abs-2303-08774},
\textbf{GPT-3.5}~\citep{NEURIPS2022_b1efde53},
\textbf{ChatGLM}~\citep{DBLP:conf/iclr/ZengLDWL0YXZXTM23,du-etal-2022-glm} and
\textbf{MedChatGLM}~\footnote{\href{https://github.com/SCIR-HI/Med-ChatGLM}{https://github.com/SCIR-HI/Med-ChatGLM}}.
For comparison, we use \textbf{BERT}~\citep{devlin-etal-2019-bert} with surpervised learning. Appendix~\ref{sec:Models} provides detailed settings.

\paragraph{Evaluations}
We extract responses from LLMs with an automatic program, and manually  check unmatched responses.
Only decisions of models (\textit{right} or \textit{wrong}) are regarded as classification results.
Unclear answers like \textit{I don't know} are neglected  in summary.

For binary classification, we evaluate performance with F1-score (F1). Additionally,  we exploit Matthews correlation coefficient (MCC) \citep{MATTHEWS1975442} to measure coefficient of predictions and labels, in order to distinguish random classifications.

\subsection{Results}

\paragraph{Probing native manipulation}
We probe LLMs with different parameters size (GPT-3.5 and ChatGLM) on simple prompts, and conduct parallel experiments on supervised BERT for comparison. 
Results are shown in Table~\ref{tab:Simple}.
\paragraph{Probing manipulation on advanced prompt}
We integrate advanced prompt to better exploit ability of LLMs, results are shown in Table~\ref{tab:all}.
This is main evidence for subsequent probing.
\definecolor{SnowyMint}{RGB}{231,248,247}
\definecolor{TitanWhite}{RGB}{226,229,243}
\begin{table}
	\centering
 \small
	\resizebox{\linewidth}{!}{
		\begin{tblr}{
			cells = {c},
			row{2} = {TitanWhite},
			cell{1}{1} = {c=2,r=2}{},
			cell{1}{3} = {c=2}{},
			cell{1}{5} = {c=2}{},
			cell{1}{7} = {c=2}{},
			cell{2}{3} = {SnowyMint},
			cell{2}{5} = {SnowyMint},
			cell{2}{7} = {SnowyMint},
			cell{3}{1} = {r=3}{},
			cell{3}{3} = {SnowyMint},
			cell{3}{4} = {TitanWhite},
			cell{3}{5} = {SnowyMint},
			cell{3}{6} = {TitanWhite},
			cell{3}{7} = {SnowyMint},
			cell{3}{8} = {TitanWhite},
			cell{4}{3} = {SnowyMint},
			cell{4}{4} = {TitanWhite},
			cell{4}{5} = {SnowyMint},
			cell{4}{6} = {TitanWhite},
			cell{4}{7} = {SnowyMint},
			cell{4}{8} = {TitanWhite},
			cell{5}{3} = {SnowyMint},
			cell{5}{4} = {TitanWhite},
			cell{5}{5} = {SnowyMint},
			cell{5}{6} = {TitanWhite},
			cell{5}{7} = {SnowyMint},
			cell{5}{8} = {TitanWhite},
			cell{6}{1} = {r=3}{},
			cell{6}{3} = {SnowyMint},
			cell{6}{4} = {TitanWhite},
			cell{6}{5} = {SnowyMint},
			cell{6}{6} = {TitanWhite},
			cell{6}{7} = {SnowyMint},
			cell{6}{8} = {TitanWhite},
			cell{7}{3} = {SnowyMint},
			cell{7}{4} = {TitanWhite},
			cell{7}{5} = {SnowyMint},
			cell{7}{6} = {TitanWhite},
			cell{7}{7} = {SnowyMint},
			cell{7}{8} = {TitanWhite},
			cell{8}{3} = {SnowyMint},
			cell{8}{4} = {TitanWhite},
			cell{8}{5} = {SnowyMint},
			cell{8}{6} = {TitanWhite},
			cell{8}{7} = {SnowyMint},
			cell{8}{8} = {TitanWhite},
			cell{9}{1} = {r=3}{},
			cell{9}{3} = {SnowyMint},
			cell{9}{4} = {TitanWhite},
			cell{9}{5} = {SnowyMint},
			cell{9}{6} = {TitanWhite},
			cell{9}{7} = {SnowyMint},
			cell{9}{8} = {TitanWhite},
			cell{10}{3} = {SnowyMint},
			cell{10}{4} = {TitanWhite},
			cell{10}{5} = {SnowyMint},
			cell{10}{6} = {TitanWhite},
			cell{10}{7} = {SnowyMint},
			cell{10}{8} = {TitanWhite},
			cell{11}{3} = {SnowyMint},
			cell{11}{4} = {TitanWhite},
			cell{11}{5} = {SnowyMint},
			cell{11}{6} = {TitanWhite},
			cell{11}{7} = {SnowyMint},
			cell{11}{8} = {TitanWhite},
			hline{1,3,6,9,12} = {-}{},
			hline{4-5,7-8,10-11} = {2-8}{dashed},
		  }
		  Models &    & ChatGLM &      & GPT-3.5 &      & BERT &      \\
		  &    & F1      & MCC  & F1      & MCC  & F1   & MCC  \\
   Act 1  & L1 & 0.63    & 0.27 & 0.67    & 0.26 & 0.79 & 0.56 \\
		  & L2 & 0.53    & 0.14 & 0.68    & 0.25 & 0.84 & 0.67 \\
		  & L3 & 0.21    & 0.06 & 0.65    & 0.11 & 0.78 & 0.52 \\
   Act 2  & L1 & 0.57    & 0.33 & 0.74    & 0.50 & 0.77 & 0.48 \\
		  & L2 & 0.52    & 0.24 & 0.73    & 0.38 & 0.80 & 0.56 \\
		  & L3 & 0.15    & 0.11 & 0.67    & 0.22 & 0.68 & 0.22 \\
   Act 3  & L1 & 0.13    & 0.04 & 0.62    & 0.24 & 0.89 & 0.76 \\
		  & L2 & 0.02    & 0.02 & 0.53    & 0.11 & 0.88 & 0.76 \\
		  & L3 & 0.16    & 0.09 & 0.52    & 0.18 & 0.81 & 0.62 
		  \end{tblr}
	}
	\caption{Overall F1 and MCC results on simple prompts, \textit{Ln} stands for knowledge enhancement in layer n.}
	\label{tab:Simple}
	\end{table}
\begin{table}
\centering
\small
\resizebox{\linewidth}{!}{
	\begin{tblr}{
		cells = {c},
		row{2} = {TitanWhite},
		cell{1}{1} = {c=2,r=2}{},
		cell{1}{3} = {c=2}{},
		cell{1}{5} = {c=2}{},
		cell{1}{7} = {c=2}{},
		cell{1}{9} = {c=2}{},
		cell{2}{3} = {SnowyMint},
		cell{2}{5} = {SnowyMint},
		cell{2}{7} = {SnowyMint},
		cell{2}{9} = {SnowyMint},
		cell{3}{1} = {r=4}{},
		cell{3}{3} = {SnowyMint},
		cell{3}{4} = {TitanWhite},
		cell{3}{5} = {SnowyMint},
		cell{3}{6} = {TitanWhite},
		cell{3}{7} = {SnowyMint},
		cell{3}{8} = {TitanWhite},
		cell{3}{9} = {SnowyMint},
		cell{3}{10} = {TitanWhite},
		cell{4}{3} = {SnowyMint},
		cell{4}{4} = {TitanWhite},
		cell{4}{5} = {SnowyMint},
		cell{4}{6} = {TitanWhite},
		cell{4}{7} = {SnowyMint},
		cell{4}{8} = {TitanWhite},
		cell{4}{9} = {SnowyMint},
		cell{4}{10} = {TitanWhite},
		cell{5}{3} = {SnowyMint},
		cell{5}{4} = {TitanWhite},
		cell{5}{5} = {SnowyMint},
		cell{5}{6} = {TitanWhite},
		cell{5}{7} = {SnowyMint},
		cell{5}{8} = {TitanWhite},
		cell{5}{9} = {SnowyMint},
		cell{5}{10} = {TitanWhite},
		cell{6}{3} = {SnowyMint},
		cell{6}{4} = {TitanWhite},
		cell{6}{5} = {SnowyMint},
		cell{6}{6} = {TitanWhite},
		cell{6}{7} = {SnowyMint},
		cell{6}{8} = {TitanWhite},
		cell{6}{9} = {SnowyMint},
		cell{6}{10} = {TitanWhite},
		cell{7}{1} = {r=4}{},
		cell{7}{3} = {SnowyMint},
		cell{7}{4} = {TitanWhite},
		cell{7}{5} = {SnowyMint},
		cell{7}{6} = {TitanWhite},
		cell{7}{7} = {SnowyMint},
		cell{7}{8} = {TitanWhite},
		cell{7}{9} = {SnowyMint},
		cell{7}{10} = {TitanWhite},
		cell{8}{3} = {SnowyMint},
		cell{8}{4} = {TitanWhite},
		cell{8}{5} = {SnowyMint},
		cell{8}{6} = {TitanWhite},
		cell{8}{7} = {SnowyMint},
		cell{8}{8} = {TitanWhite},
		cell{8}{9} = {SnowyMint},
		cell{8}{10} = {TitanWhite},
		cell{9}{3} = {SnowyMint},
		cell{9}{4} = {TitanWhite},
		cell{9}{5} = {SnowyMint},
		cell{9}{6} = {TitanWhite},
		cell{9}{7} = {SnowyMint},
		cell{9}{8} = {TitanWhite},
		cell{9}{9} = {SnowyMint},
		cell{9}{10} = {TitanWhite},
		cell{10}{3} = {SnowyMint},
		cell{10}{4} = {TitanWhite},
		cell{10}{5} = {SnowyMint},
		cell{10}{6} = {TitanWhite},
		cell{10}{7} = {SnowyMint},
		cell{10}{8} = {TitanWhite},
		cell{10}{9} = {SnowyMint},
		cell{10}{10} = {TitanWhite},
		cell{11}{1} = {r=4}{},
		cell{11}{3} = {SnowyMint},
		cell{11}{4} = {TitanWhite},
		cell{11}{5} = {SnowyMint},
		cell{11}{6} = {TitanWhite},
		cell{11}{7} = {SnowyMint},
		cell{11}{8} = {TitanWhite},
		cell{11}{9} = {SnowyMint},
		cell{11}{10} = {TitanWhite},
		cell{12}{3} = {SnowyMint},
		cell{12}{4} = {TitanWhite},
		cell{12}{5} = {SnowyMint},
		cell{12}{6} = {TitanWhite},
		cell{12}{7} = {SnowyMint},
		cell{12}{8} = {TitanWhite},
		cell{12}{9} = {SnowyMint},
		cell{12}{10} = {TitanWhite},
		cell{13}{3} = {SnowyMint},
		cell{13}{4} = {TitanWhite},
		cell{13}{5} = {SnowyMint},
		cell{13}{6} = {TitanWhite},
		cell{13}{7} = {SnowyMint},
		cell{13}{8} = {TitanWhite},
		cell{13}{9} = {SnowyMint},
		cell{13}{10} = {TitanWhite},
		cell{14}{3} = {SnowyMint},
		cell{14}{4} = {TitanWhite},
		cell{14}{5} = {SnowyMint},
		cell{14}{6} = {TitanWhite},
		cell{14}{7} = {SnowyMint},
		cell{14}{8} = {TitanWhite},
		cell{14}{9} = {SnowyMint},
		cell{14}{10} = {TitanWhite},
		hline{1,3,11,15} = {-}{},
		hline{4-6,8-10,12-14} = {2-10}{dashed},
		hline{7} = {1}{dashed},
		hline{7} = {2-10}{},
	  }
	  Models &      & GPT 4 &      & GPT-3.5 &      & ChatGLM &      & MedChatGLM &       \\
			 &      & F1    & MCC  & F1      & MCC  & F1      & MCC  & F1     & MCC   \\
	  Act 1  & L1   & 0.71  & 0.33 & 0.68    & 0.27 & 0.63    & 0.14 & 0.52   & 0.06  \\
			 & L2   & 0.86  & 0.71 & 0.75    & 0.42 & 0.53    & 0.21 & 0.57   & 0.08  \\
			 & L2.5 & 0.81  & 0.60 & 0.16    & 0.13 & 0.47    & 0.13 & 0.55   & 0.14  \\
			 & L3   & 0.76  & 0.46 & 0.68    & 0.20 & 0.26    & 0.10 & 0.41   & -0.10 \\
	  Act 2  & L1   & 0.75  & 0.45 & 0.70    & 0.37 & 0.63    & 0.20 & 0.57   & 0.14  \\
			 & L2   & 0.84  & 0.66 & 0.80    & 0.56 & 0.50    & 0.30 & 0.55   & -0.02 \\
			 & L2.5 & 0.79  & 0.55 & 0.76    & 0.46 & 0.63    & 0.40 & 0.58   & 0.08  \\
			 & L3   & 0.75  & 0.46 & 0.73    & 0.36 & 0.21    & 0.12 & 0.52   & -0.04 \\
	  Act 3  & L1   & 0.41  & 0.39 & 0.50    & 0.21 & 0.26    & 0.07 & 0.41   & -0.04 \\
			 & L2   & 0.39  & 0.37 & 0.34    & 0.12 & 0.06    & 0.00 & 0.51   & -0.04 \\
			 & L2.5 & 0.60  & 0.50 & 0.36    & 0.17 & 0.07    & 0.01 & 0.30   & -0.23 \\
			 & L3   & 0.58  & 0.48 & 0.42    & 0.19 & 0.04    & 0.03 & 0.50    & 0.04  
	  \end{tblr}
}
\caption{Overall results for advanced prompts, \textit{Ln} stands for knowledge enhancement in layer n.}
\label{tab:all}
\end{table}
 \subsection{Analysis}
\begin{figure}[!ht]
	\begin{center}
		\includegraphics[width=1.0\columnwidth]{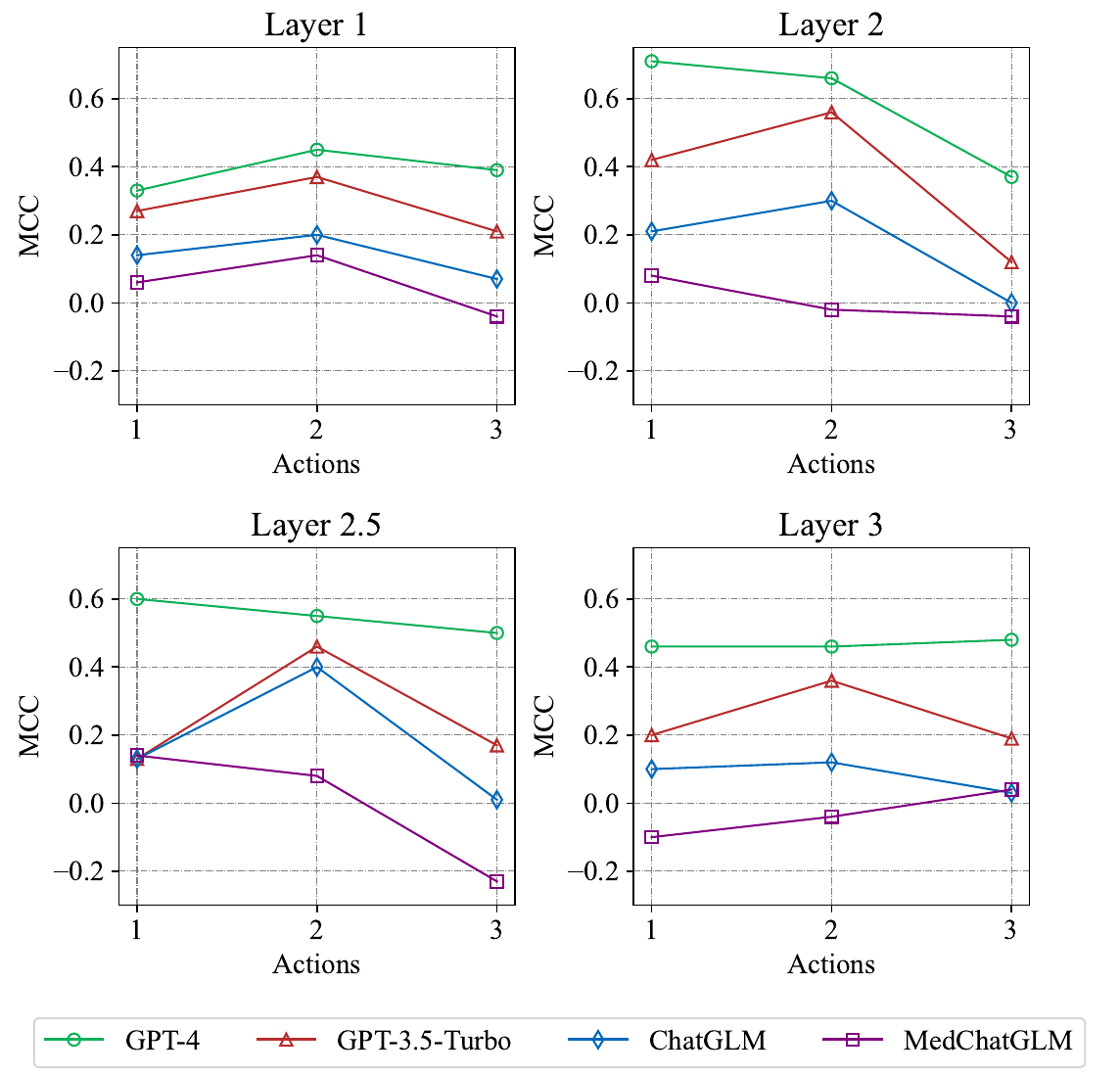}
		\caption{Trend of MCC with three actions in different models and different layers (Using Advanced Prompt).}
		\label{fig:preferences}
	\end{center}
\end{figure}

\paragraph{Overall Analysis}
(1) Experimental results of MCC show that LLMs have weak causality ability on given classification task.
But it is not comparable with supervised models like BERT.
(2) Performance of LLMs varies. Reasons may include parameters, training strategies and domain knowledge. We discuss this in Appendix~\ref{ModelSelection}.
(3) Additionally, models show preferences for different actions in dataset, as shown in Fig.~\ref{fig:preferences}.

\paragraph{Global Semantics}

Global semantic is the key for classification, we derive this from actions preferences of models.
Action 2 integrates more modification from statistical perspective, which is easy for models to distinguish.
We evaluate perplexity of sentences in Appendix.~\ref{sec:ppl} to prove this.
GPT-4 performs better in action 1 when knowledge is given, since it gains more instruction ability.
This tendency excludes MedChatGLM, as its MCC approaches to 0 and not indicative for analysis.
This tendency persists regardless of the prompts used.
\paragraph{Entities In Causality}
(1) Discovering entities in causality is less important in manipulation than global semantics.
This is derived from action preferences as mentioned above. Moreover, we conduct back-translation experiments for augmented knowledge in layer 2 and observe a performance drop for most cases.
Back-translation tends to preserve entities in sentences, since expressions of medical entities are usually standard in Chinese and English.
(2) LLMs lack ability of aligning entities and their establishing systematical relations.
Entities in sentence are focused as a result of attention mechanism, but LLMs except for GPT-4 can not exploit augmentation of layer 3.
Layer 3 should be beneficial in Appendix.~\ref{sec:Effectiveness_Layer3}. Comparing with layer 2.5, layer 3 provides entities in other form, and this indicates LLMs fail to recognize causes and effects in heterogeneous form.

\paragraph{Causality Cognition}
LLMs do not show much specific cognition for causality, relying more on linguistic order and positions to demonstrate causality. 
The performance of action 3 is the worst of the three and has even approached random categorization for some models (ChatGLM and MedChatGLM).
Augmentation of layer 1 even causes a performance drop, regardless prompt used for all models including GPT-4.
In contrast, the introduction of layer 2 in action 1 improves performance.
This is because action 3 disturbs mutual causation. To realize mutual disturbance (especially when gold standard is given in layer 1), models needs to recognize causation specifically first.

\paragraph{Knowledge}
Background knowledge has little contributions, and only augmentation of layer 2 assist for classifications after guidance of advanced prompt.
(1) LLMs are native to believe its pretrained knowledge for classification. 
(2) Background knowledge about causality confuses LLMs for classification, and intervene normal manipulation by internal knowledge.
(3) LLMs manipulation of causality relies on abstract principles summarized from internal knowledge, which is in an abstract level. Since MedChatGLM diminishes classification abilities of ChatGLM, and approaches to random classification.

\section{Conclusion}
In this paper, we introduce an innovative structure tailored to investigate intrinsic manipulations of causality for LLMs. 
We construct a classification dataset focusing on causal relations and entities in sentences.
Then we probe models' performance on this classification dataset.
We provide "shortcuts" through RAG and ICL, and observe performance change in datasets.
Probing conclusion is derived by judging whether such shortcuts are beneficial.
Our result indicates that LLMs show certain ability of causal recognition, mainly as a result of global semantic. Causal entities and their relations lack for detailed and specific manipulation, especially for LLMs with smaller parameters.
Our probing work still has limitations. (1) Our conclusion is derived as a summary for various LLMs. Relation of causality and LLMs' training strategies should be discussed.
(2) Our experiment lacks the ability for detailed discussion about supervised learning and zero-shot cases for causality.
\bibliography{Cattail_cite_vol1}

\begin{thebibliography}{36}
\providecommand{\natexlab}[1]{#1}

\bibitem[{Bareinboim et~al.(2022)Bareinboim, Correa, Ibeling, and Icard}]{DBLP:books/acm/22/BareinboimCII22}
Elias Bareinboim, Juan~D. Correa, Duligur Ibeling, and Thomas Icard. 2022.
\newblock \href {https://doi.org/10.1145/3501714.3501743} {On pearl's hierarchy and the foundations of causal inference}.
\newblock In Hector Geffner, Rina Dechter, and Joseph~Y. Halpern, editors, \emph{Probabilistic and Causal Inference: The Works of Judea Pearl}, volume~36 of \emph{{ACM} Books}, pages 507--556. {ACM}.

\bibitem[{Bhagavatula et~al.(2020)Bhagavatula, Bras, Malaviya, Sakaguchi, Holtzman, Rashkin, Downey, Yih, and Choi}]{DBLP:conf/iclr/BhagavatulaBMSH20}
Chandra Bhagavatula, Ronan~Le Bras, Chaitanya Malaviya, Keisuke Sakaguchi, Ari Holtzman, Hannah Rashkin, Doug Downey, Wen{-}tau Yih, and Yejin Choi. 2020.
\newblock \href {https://openreview.net/forum?id=Byg1v1HKDB} {Abductive commonsense reasoning}.
\newblock In \emph{8th International Conference on Learning Representations, {ICLR} 2020, Addis Ababa, Ethiopia, April 26-30, 2020}. OpenReview.net.

\bibitem[{Brown et~al.(2020)Brown, Mann, Ryder, Subbiah, Kaplan, Dhariwal, Neelakantan, Shyam, Sastry, Askell, Agarwal, Herbert-Voss, Krueger, Henighan, Child, Ramesh, Ziegler, Wu, Winter, Hesse, Chen, Sigler, Litwin, Gray, Chess, Clark, Berner, McCandlish, Radford, Sutskever, and Amodei}]{brown2020language}
Tom~B. Brown, Benjamin Mann, Nick Ryder, Melanie Subbiah, Jared Kaplan, Prafulla Dhariwal, Arvind Neelakantan, Pranav Shyam, Girish Sastry, Amanda Askell, Sandhini Agarwal, Ariel Herbert-Voss, Gretchen Krueger, Tom Henighan, Rewon Child, Aditya Ramesh, Daniel~M. Ziegler, Jeffrey Wu, Clemens Winter, Christopher Hesse, Mark Chen, Eric Sigler, Mateusz Litwin, Scott Gray, Benjamin Chess, Jack Clark, Christopher Berner, Sam McCandlish, Alec Radford, Ilya Sutskever, and Dario Amodei. 2020.
\newblock Language models are few-shot learners.
\newblock In \emph{Proceedings of the 34th International Conference on Neural Information Processing Systems}, NIPS'20, Red Hook, NY, USA. Curran Associates Inc.

\bibitem[{Chen et~al.(2017)Chen, Fisch, Weston, and Bordes}]{chen-etal-2017-reading}
Danqi Chen, Adam Fisch, Jason Weston, and Antoine Bordes. 2017.
\newblock \href {https://doi.org/10.18653/v1/P17-1171} {Reading {W}ikipedia to answer open-domain questions}.
\newblock In \emph{Proceedings of the 55th Annual Meeting of the Association for Computational Linguistics (Volume 1: Long Papers)}, pages 1870--1879, Vancouver, Canada. Association for Computational Linguistics.

\bibitem[{Chen et~al.(2023)Chen, Shi, Fu, Cheng, Li, and Xiao}]{chen-etal-2023-say}
Jiangjie Chen, Wei Shi, Ziquan Fu, Sijie Cheng, Lei Li, and Yanghua Xiao. 2023.
\newblock \href {https://doi.org/10.18653/v1/2023.acl-long.550} {Say what you mean! large language models speak too positively about negative commonsense knowledge}.
\newblock In \emph{Proceedings of the 61st Annual Meeting of the Association for Computational Linguistics (Volume 1: Long Papers)}, pages 9890--9908, Toronto, Canada. Association for Computational Linguistics.

\bibitem[{DeepSeek-AI et~al.(2024)DeepSeek-AI, :, Bi, Chen, Chen, Chen, Dai, Deng, Ding, Dong, Du, Fu, Gao, Gao, Gao, Ge, Guan, Guo, Guo, Hao, Hao, He, Hu, Huang, Li, Li, Li, Li, Li, Liang, Lin, Liu, Liu, Liu, Liu, Liu, Liu, Lu, Lu, Luo, Ma, Nie, Pei, Piao, Qiu, Qu, Ren, Ren, Ruan, Sha, Shao, Song, Su, Sun, Sun, Tang, Wang, Wang, Wang, Wang, Wang, Wu, Wu, Xie, Xie, Xie, Xiong, Xu, Xu, Xu, Yang, You, Yu, Yu, Zhang, Zhang, Zhang, Zhang, Zhang, Zhang, Zhang, Zhang, Zhao, Zhao, Zhou, Zhou, Zhu, and Zou}]{deepseekai2024deepseek}
DeepSeek-AI, :, Xiao Bi, Deli Chen, Guanting Chen, Shanhuang Chen, Damai Dai, Chengqi Deng, Honghui Ding, Kai Dong, Qiushi Du, Zhe Fu, Huazuo Gao, Kaige Gao, Wenjun Gao, Ruiqi Ge, Kang Guan, Daya Guo, Jianzhong Guo, Guangbo Hao, Zhewen Hao, Ying He, Wenjie Hu, Panpan Huang, Erhang Li, Guowei Li, Jiashi Li, Yao Li, Y.~K. Li, Wenfeng Liang, Fangyun Lin, A.~X. Liu, Bo~Liu, Wen Liu, Xiaodong Liu, Xin Liu, Yiyuan Liu, Haoyu Lu, Shanghao Lu, Fuli Luo, Shirong Ma, Xiaotao Nie, Tian Pei, Yishi Piao, Junjie Qiu, Hui Qu, Tongzheng Ren, Zehui Ren, Chong Ruan, Zhangli Sha, Zhihong Shao, Junxiao Song, Xuecheng Su, Jingxiang Sun, Yaofeng Sun, Minghui Tang, Bingxuan Wang, Peiyi Wang, Shiyu Wang, Yaohui Wang, Yongji Wang, Tong Wu, Y.~Wu, Xin Xie, Zhenda Xie, Ziwei Xie, Yiliang Xiong, Hanwei Xu, R.~X. Xu, Yanhong Xu, Dejian Yang, Yuxiang You, Shuiping Yu, Xingkai Yu, B.~Zhang, Haowei Zhang, Lecong Zhang, Liyue Zhang, Mingchuan Zhang, Minghua Zhang, Wentao Zhang, Yichao Zhang, Chenggang Zhao, Yao Zhao, Shangyan Zhou, Shunfeng Zhou, Qihao Zhu, and Yuheng Zou. 2024.
\newblock \href {https://arxiv.org/abs/2401.02954} {Deepseek llm: Scaling open-source language models with longtermism}.
\newblock \emph{Preprint}, arXiv:2401.02954.

\bibitem[{Devlin et~al.(2019)Devlin, Chang, Lee, and Toutanova}]{devlin-etal-2019-bert}
Jacob Devlin, Ming-Wei Chang, Kenton Lee, and Kristina Toutanova. 2019.
\newblock \href {https://doi.org/10.18653/v1/N19-1423} {{BERT}: Pre-training of deep bidirectional transformers for language understanding}.
\newblock In \emph{Proceedings of the 2019 Conference of the North {A}merican Chapter of the Association for Computational Linguistics: Human Language Technologies, Volume 1 (Long and Short Papers)}, pages 4171--4186, Minneapolis, Minnesota. Association for Computational Linguistics.

\bibitem[{Du et~al.(2022{\natexlab{a}})Du, Ding, Xiong, Liu, and Qin}]{du-etal-2022-e}
Li~Du, Xiao Ding, Kai Xiong, Ting Liu, and Bing Qin. 2022{\natexlab{a}}.
\newblock \href {https://doi.org/10.18653/v1/2022.acl-long.33} {e-{CARE}: a new dataset for exploring explainable causal reasoning}.
\newblock In \emph{Proceedings of the 60th Annual Meeting of the Association for Computational Linguistics (Volume 1: Long Papers)}, pages 432--446, Dublin, Ireland. Association for Computational Linguistics.

\bibitem[{Du et~al.(2022{\natexlab{b}})Du, Qian, Liu, Ding, Qiu, Yang, and Tang}]{du-etal-2022-glm}
Zhengxiao Du, Yujie Qian, Xiao Liu, Ming Ding, Jiezhong Qiu, Zhilin Yang, and Jie Tang. 2022{\natexlab{b}}.
\newblock \href {https://doi.org/10.18653/v1/2022.acl-long.26} {{GLM}: General language model pretraining with autoregressive blank infilling}.
\newblock In \emph{Proceedings of the 60th Annual Meeting of the Association for Computational Linguistics (Volume 1: Long Papers)}, pages 320--335, Dublin, Ireland. Association for Computational Linguistics.

\bibitem[{Ganguli et~al.(2022)Ganguli, Hernandez, Lovitt, Askell, Bai, Chen, Conerly, DasSarma, Drain, Elhage, Showk, Fort, Hatfield{-}Dodds, Henighan, Johnston, Jones, Joseph, Kernian, Kravec, Mann, Nanda, Ndousse, Olsson, Amodei, Brown, Kaplan, McCandlish, Olah, Amodei, and Clark}]{DBLP:conf/fat/GanguliHLABCCDD22}
Deep Ganguli, Danny Hernandez, Liane Lovitt, Amanda Askell, Yuntao Bai, Anna Chen, Tom Conerly, Nova DasSarma, Dawn Drain, Nelson Elhage, Sheer~El Showk, Stanislav Fort, Zac Hatfield{-}Dodds, Tom Henighan, Scott Johnston, Andy Jones, Nicholas Joseph, Jackson Kernian, Shauna Kravec, Ben Mann, Neel Nanda, Kamal Ndousse, Catherine Olsson, Daniela Amodei, Tom Brown, Jared Kaplan, Sam McCandlish, Christopher Olah, Dario Amodei, and Jack Clark. 2022.
\newblock \href {https://doi.org/10.1145/3531146.3533229} {Predictability and surprise in large generative models}.
\newblock In \emph{FAccT '22: 2022 {ACM} Conference on Fairness, Accountability, and Transparency, Seoul, Republic of Korea, June 21 - 24, 2022}, pages 1747--1764. {ACM}.

\bibitem[{Gao et~al.(2023)Gao, Ding, Qin, and Liu}]{gao-etal-2023-chatgpt}
Jinglong Gao, Xiao Ding, Bing Qin, and Ting Liu. 2023.
\newblock \href {https://doi.org/10.18653/v1/2023.findings-emnlp.743} {Is {C}hat{GPT} a good causal reasoner? a comprehensive evaluation}.
\newblock In \emph{Findings of the Association for Computational Linguistics: EMNLP 2023}, pages 11111--11126, Singapore. Association for Computational Linguistics.

\bibitem[{Gao et~al.(2019)Gao, Choubey, and Huang}]{gao-etal-2019-modeling}
Lei Gao, Prafulla~Kumar Choubey, and Ruihong Huang. 2019.
\newblock \href {https://doi.org/10.18653/v1/N19-1179} {Modeling document-level causal structures for event causal relation identification}.
\newblock In \emph{Proceedings of the 2019 Conference of the North {A}merican Chapter of the Association for Computational Linguistics: Human Language Technologies, Volume 1 (Long and Short Papers)}, pages 1808--1817, Minneapolis, Minnesota. Association for Computational Linguistics.

\bibitem[{Hewitt and Manning(2019)}]{hewitt-manning-2019-structural}
John Hewitt and Christopher~D. Manning. 2019.
\newblock \href {https://doi.org/10.18653/v1/N19-1419} {{A} structural probe for finding syntax in word representations}.
\newblock In \emph{Proceedings of the 2019 Conference of the North {A}merican Chapter of the Association for Computational Linguistics: Human Language Technologies, Volume 1 (Long and Short Papers)}, pages 4129--4138, Minneapolis, Minnesota. Association for Computational Linguistics.

\bibitem[{Hossain et~al.(2023)Hossain, Dev, and Singh}]{hossain-etal-2023-misgendered}
Tamanna Hossain, Sunipa Dev, and Sameer Singh. 2023.
\newblock \href {https://doi.org/10.18653/v1/2023.acl-long.293} {{MISGENDERED}: Limits of large language models in understanding pronouns}.
\newblock In \emph{Proceedings of the 61st Annual Meeting of the Association for Computational Linguistics (Volume 1: Long Papers)}, pages 5352--5367, Toronto, Canada. Association for Computational Linguistics.

\bibitem[{Jiang et~al.(2023)Jiang, Sablayrolles, Mensch, Bamford, Chaplot, de~las Casas, Bressand, Lengyel, Lample, Saulnier, Lavaud, Lachaux, Stock, Scao, Lavril, Wang, Lacroix, and Sayed}]{jiang2023mistral}
Albert~Q. Jiang, Alexandre Sablayrolles, Arthur Mensch, Chris Bamford, Devendra~Singh Chaplot, Diego de~las Casas, Florian Bressand, Gianna Lengyel, Guillaume Lample, Lucile Saulnier, Lélio~Renard Lavaud, Marie-Anne Lachaux, Pierre Stock, Teven~Le Scao, Thibaut Lavril, Thomas Wang, Timothée Lacroix, and William~El Sayed. 2023.
\newblock \href {https://arxiv.org/abs/2310.06825} {Mistral 7b}.
\newblock \emph{Preprint}, arXiv:2310.06825.

\bibitem[{Jin et~al.(2024)Jin, Liu, LYU, Poff, Sachan, Mihalcea, Diab, and Sch{\"o}lkopf}]{jin2024can}
Zhijing Jin, Jiarui Liu, Zhiheng LYU, Spencer Poff, Mrinmaya Sachan, Rada Mihalcea, Mona~T. Diab, and Bernhard Sch{\"o}lkopf. 2024.
\newblock \href {https://openreview.net/forum?id=vqIH0ObdqL} {Can large language models infer causation from correlation?}
\newblock In \emph{The Twelfth International Conference on Learning Representations}.

\bibitem[{Johnson et~al.(2021)Johnson, Douze, and J{\'e}gou}]{8733051}
Jeff Johnson, Matthijs Douze, and Herv{\'e} J{\'e}gou. 2021.
\newblock \href {https://doi.org/10.1109/TBDATA.2019.2921572} {Billion-scale similarity search with gpus}.
\newblock \emph{IEEE Transactions on Big Data}, 7(3):535--547.

\bibitem[{Kiciman et~al.(2023)Kiciman, Ness, Sharma, and Tan}]{CausalFrontier}
Emre Kiciman, Robert Ness, Amit Sharma, and Chenhao Tan. 2023.
\newblock \href {https://doi.org/10.48550/ARXIV.2305.00050} {Causal reasoning and large language models: Opening a new frontier for causality}.
\newblock \emph{CoRR}, abs/2305.00050.

\bibitem[{Kojima et~al.(2022)Kojima, Gu, Reid, Matsuo, and Iwasawa}]{DBLP:conf/nips/KojimaGRMI22}
Takeshi Kojima, Shixiang~Shane Gu, Machel Reid, Yutaka Matsuo, and Yusuke Iwasawa. 2022.
\newblock \href {http://papers.nips.cc/paper\_files/paper/2022/hash/8bb0d291acd4acf06ef112099c16f326-Abstract-Conference.html} {Large language models are zero-shot reasoners}.
\newblock In \emph{NeurIPS}.

\bibitem[{Lin et~al.(2020)Lin, Zhou, Shen, Zhou, Bhagavatula, Choi, and Ren}]{lin-etal-2020-commongen}
Bill~Yuchen Lin, Wangchunshu Zhou, Ming Shen, Pei Zhou, Chandra Bhagavatula, Yejin Choi, and Xiang Ren. 2020.
\newblock \href {https://doi.org/10.18653/v1/2020.findings-emnlp.165} {{C}ommon{G}en: A constrained text generation challenge for generative commonsense reasoning}.
\newblock In \emph{Findings of the Association for Computational Linguistics: EMNLP 2020}, pages 1823--1840, Online. Association for Computational Linguistics.

\bibitem[{Matthews(1975)}]{MATTHEWS1975442}
B.W. Matthews. 1975.
\newblock \href {https://doi.org/10.1016/0005-2795(75)90109-9} {Comparison of the predicted and observed secondary structure of t4 phage lysozyme}.
\newblock \emph{Biochimica et Biophysica Acta (BBA) - Protein Structure}, 405(2):442--451.

\bibitem[{Mu and Li(2023)}]{mu-li-2023-enhancing}
Feiteng Mu and Wenjie Li. 2023.
\newblock \href {https://doi.org/10.18653/v1/2023.acl-short.83} {Enhancing event causality identification with counterfactual reasoning}.
\newblock In \emph{Proceedings of the 61st Annual Meeting of the Association for Computational Linguistics (Volume 2: Short Papers)}, pages 967--975, Toronto, Canada. Association for Computational Linguistics.

\bibitem[{OpenAI(2023)}]{DBLP:journals/corr/abs-2303-08774}
OpenAI. 2023.
\newblock \href {https://doi.org/10.48550/arXiv.2303.08774} {{GPT-4} technical report}.
\newblock \emph{CoRR}, abs/2303.08774.

\bibitem[{Ouyang et~al.(2022)Ouyang, Wu, Jiang, Almeida, Wainwright, Mishkin, Zhang, Agarwal, Slama, Ray, Schulman, Hilton, Kelton, Miller, Simens, Askell, Welinder, Christiano, Leike, and Lowe}]{NEURIPS2022_b1efde53}
Long Ouyang, Jeffrey Wu, Xu~Jiang, Diogo Almeida, Carroll Wainwright, Pamela Mishkin, Chong Zhang, Sandhini Agarwal, Katarina Slama, Alex Ray, John Schulman, Jacob Hilton, Fraser Kelton, Luke Miller, Maddie Simens, Amanda Askell, Peter Welinder, Paul~F Christiano, Jan Leike, and Ryan Lowe. 2022.
\newblock \href {https://proceedings.neurips.cc/paper_files/paper/2022/file/b1efde53be364a73914f58805a001731-Paper-Conference.pdf} {Training language models to follow instructions with human feedback}.
\newblock In \emph{Advances in Neural Information Processing Systems}, volume~35, pages 27730--27744. Curran Associates, Inc.

\bibitem[{Radford et~al.(2019)Radford, Wu, Child, Luan, Amodei, and Sutskever}]{radford2019language}
Alec Radford, Jeff Wu, Rewon Child, David Luan, Dario Amodei, and Ilya Sutskever. 2019.
\newblock Language models are unsupervised multitask learners.

\bibitem[{Ramezani and Xu(2023)}]{ramezani-xu-2023-knowledge}
Aida Ramezani and Yang Xu. 2023.
\newblock \href {https://doi.org/10.18653/v1/2023.acl-long.26} {Knowledge of cultural moral norms in large language models}.
\newblock In \emph{Proceedings of the 61st Annual Meeting of the Association for Computational Linguistics (Volume 1: Long Papers)}, pages 428--446, Toronto, Canada. Association for Computational Linguistics.

\bibitem[{Reimers and Gurevych(2019)}]{reimers-gurevych-2019-sentence}
Nils Reimers and Iryna Gurevych. 2019.
\newblock \href {https://doi.org/10.18653/v1/D19-1410} {Sentence-{BERT}: Sentence embeddings using {S}iamese {BERT}-networks}.
\newblock In \emph{Proceedings of the 2019 Conference on Empirical Methods in Natural Language Processing and the 9th International Joint Conference on Natural Language Processing (EMNLP-IJCNLP)}, pages 3982--3992, Hong Kong, China. Association for Computational Linguistics.

\bibitem[{Stolfo et~al.(2023)Stolfo, Jin, Shridhar, Schoelkopf, and Sachan}]{stolfo-etal-2023-causal}
Alessandro Stolfo, Zhijing Jin, Kumar Shridhar, Bernhard Schoelkopf, and Mrinmaya Sachan. 2023.
\newblock \href {https://doi.org/10.18653/v1/2023.acl-long.32} {A causal framework to quantify the robustness of mathematical reasoning with language models}.
\newblock In \emph{Proceedings of the 61st Annual Meeting of the Association for Computational Linguistics (Volume 1: Long Papers)}, pages 545--561, Toronto, Canada. Association for Computational Linguistics.

\bibitem[{Talmor et~al.(2019)Talmor, Herzig, Lourie, and Berant}]{talmor-etal-2019-commonsenseqa}
Alon Talmor, Jonathan Herzig, Nicholas Lourie, and Jonathan Berant. 2019.
\newblock \href {https://doi.org/10.18653/v1/N19-1421} {{C}ommonsense{QA}: A question answering challenge targeting commonsense knowledge}.
\newblock In \emph{Proceedings of the 2019 Conference of the North {A}merican Chapter of the Association for Computational Linguistics: Human Language Technologies, Volume 1 (Long and Short Papers)}, pages 4149--4158, Minneapolis, Minnesota. Association for Computational Linguistics.

\bibitem[{Tan et~al.(2019)Tan, Yu, and Bansal}]{tan-etal-2019-learning}
Hao Tan, Licheng Yu, and Mohit Bansal. 2019.
\newblock \href {https://doi.org/10.18653/v1/N19-1268} {Learning to navigate unseen environments: Back translation with environmental dropout}.
\newblock In \emph{Proceedings of the 2019 Conference of the North {A}merican Chapter of the Association for Computational Linguistics: Human Language Technologies, Volume 1 (Long and Short Papers)}, pages 2610--2621, Minneapolis, Minnesota. Association for Computational Linguistics.

\bibitem[{Wei et~al.(2022{\natexlab{a}})Wei, Tay, Bommasani, Raffel, Zoph, Borgeaud, Yogatama, Bosma, Zhou, Metzler, Chi, Hashimoto, Vinyals, Liang, Dean, and Fedus}]{wei2022emergent}
Jason Wei, Yi~Tay, Rishi Bommasani, Colin Raffel, Barret Zoph, Sebastian Borgeaud, Dani Yogatama, Maarten Bosma, Denny Zhou, Donald Metzler, Ed~H. Chi, Tatsunori Hashimoto, Oriol Vinyals, Percy Liang, Jeff Dean, and William Fedus. 2022{\natexlab{a}}.
\newblock \href {https://openreview.net/forum?id=yzkSU5zdwD} {Emergent abilities of large language models}.
\newblock \emph{Trans. Mach. Learn. Res.}, 2022.

\bibitem[{Wei et~al.(2022{\natexlab{b}})Wei, Wang, Schuurmans, Bosma, Ichter, Xia, Chi, Le, and Zhou}]{DBLP:conf/nips/Wei0SBIXCLZ22}
Jason Wei, Xuezhi Wang, Dale Schuurmans, Maarten Bosma, Brian Ichter, Fei Xia, Ed~H. Chi, Quoc~V. Le, and Denny Zhou. 2022{\natexlab{b}}.
\newblock \href {http://papers.nips.cc/paper\_files/paper/2022/hash/9d5609613524ecf4f15af0f7b31abca4-Abstract-Conference.html} {Chain-of-thought prompting elicits reasoning in large language models}.
\newblock In \emph{NeurIPS}.

\bibitem[{Ze{\v{c}}evi{\'c} et~al.(2023)Ze{\v{c}}evi{\'c}, Willig, Dhami, and Kersting}]{CausalParrots}
Matej Ze{\v{c}}evi{\'c}, Moritz Willig, Devendra~Singh Dhami, and Kristian Kersting. 2023.
\newblock \href {https://openreview.net/forum?id=tv46tCzs83} {Causal parrots: Large language models may talk causality but are not causal}.
\newblock \emph{Transactions on Machine Learning Research}.

\bibitem[{Zeng et~al.(2023)Zeng, Liu, Du, Wang, Lai, Ding, Yang, Xu, Zheng, Xia, Tam, Ma, Xue, Zhai, Chen, Liu, Zhang, Dong, and Tang}]{DBLP:conf/iclr/ZengLDWL0YXZXTM23}
Aohan Zeng, Xiao Liu, Zhengxiao Du, Zihan Wang, Hanyu Lai, Ming Ding, Zhuoyi Yang, Yifan Xu, Wendi Zheng, Xiao Xia, Weng~Lam Tam, Zixuan Ma, Yufei Xue, Jidong Zhai, Wenguang Chen, Zhiyuan Liu, Peng Zhang, Yuxiao Dong, and Jie Tang. 2023.
\newblock \href {https://openreview.net/pdf?id=-Aw0rrrPUF} {{GLM-130B:} an open bilingual pre-trained model}.
\newblock In \emph{The Eleventh International Conference on Learning Representations, {ICLR} 2023, Kigali, Rwanda, May 1-5, 2023}. OpenReview.net.

\bibitem[{Zhang et~al.(2022)Zhang, Chen, Bi, Liang, Li, Shang, Yin, Tan, Xu, Huang, Si, Ni, Xie, Sui, Chang, Zong, Yuan, Li, Yan, Zan, Zhang, Tang, and Chen}]{zhang-etal-2022-cblue}
Ningyu Zhang, Mosha Chen, Zhen Bi, Xiaozhuan Liang, Lei Li, Xin Shang, Kangping Yin, Chuanqi Tan, Jian Xu, Fei Huang, Luo Si, Yuan Ni, Guotong Xie, Zhifang Sui, Baobao Chang, Hui Zong, Zheng Yuan, Linfeng Li, Jun Yan, Hongying Zan, Kunli Zhang, Buzhou Tang, and Qingcai Chen. 2022.
\newblock \href {https://doi.org/10.18653/v1/2022.acl-long.544} {{CBLUE}: A {C}hinese biomedical language understanding evaluation benchmark}.
\newblock In \emph{Proceedings of the 60th Annual Meeting of the Association for Computational Linguistics (Volume 1: Long Papers)}, pages 7888--7915, Dublin, Ireland. Association for Computational Linguistics.

\bibitem[{Zhou et~al.(2023)Zhou, Sch{\"{a}}rli, Hou, Wei, Scales, Wang, Schuurmans, Cui, Bousquet, Le, and Chi}]{DBLP:conf/iclr/ZhouSHWS0SCBLC23}
Denny Zhou, Nathanael Sch{\"{a}}rli, Le~Hou, Jason Wei, Nathan Scales, Xuezhi Wang, Dale Schuurmans, Claire Cui, Olivier Bousquet, Quoc~V. Le, and Ed~H. Chi. 2023.
\newblock \href {https://openreview.net/pdf?id=WZH7099tgfM} {Least-to-most prompting enables complex reasoning in large language models}.
\newblock In \emph{The Eleventh International Conference on Learning Representations, {ICLR} 2023, Kigali, Rwanda, May 1-5, 2023}. OpenReview.net.

\end{thebibliography}
\clearpage
\appendix
\section{Datasets Details And Statistics}\label{sec:Apd_dataset_details}
The CMedCausal dataset~\citep{zhang-etal-2022-cblue} defines three key types of medical causal reasoning relationships: \textbf{causation}, \textbf{conditionality}, and \textbf{hierarchical relationships}, consisting of 9,153 segments of medical text and 79,244 pairs of entity relationships. Our work primarily discusses relationships related to causality, hence, we have discarded hierarchical relationships. Medical concept fragments in the dataset refer to continuous character segments that can act as independent semantic units. These segments may represent medical entities, clinical findings, or specific disease symptoms. From the perspective of expressing causal predicates, these fragments fulfill semantic roles of conditions, causes, or consequences. 

We translated part of the content from the Chinese dataset into English as an example. For instance, \textit{"Gastrointestinal dysfunction in the human body leads to a decrease in the patient's absorption capacity."} In this case, \textit{"gastrointestinal dysfunction"} is a medical concept fragment. \textit{"gastrointestinal dysfunction"} is a direct cause of \textit{"decreased absorption capacity"}, and \textit{"decreased absorption capacity"} is a direct result of \textit{"gastrointestinal dysfunction"}. We can label this data as <\textit{"gastrointestinal dysfunction"}, \textit{"decreased absorption capacity"}, \textit{"causation"}>. Here, \textit{"gastrointestinal dysfunction"} serves as the subject (\textbf{Head}) of the relationship, \textit{"decreased absorption capacity"} as the object (\textbf{Tail}), and \textit{"causation"} as the specific type of relation (\textbf{Relation}). 

In the dataset, all data can be annotated in the triplet form of <Head, Tail, Relation>. We have conducted statistical analysis on the average length of data and the number of triplets corresponding to each relation in the dataset. The specific statistical results can be found in Table~\ref{tab:statistical}.

Moreover, CMedCausal is in Chinese, and Chinese phrases contain fewer variations.
So that it is feasible to modify original dataset with text substitution, and preserve sentences fluency.

\begin{table}
    \centering
    \small

    \begin{tabular}{l||c|c|c|c|} 
    \hline
    Items                                  & Sum  & Avg & Max & Min  \\ 
    \hline
    Passages                               & 999  & -   & -   & -    \\
    Length of each passage & -    & 267 & 544 & 29   \\
    Relations per instance                 & 8804 & 8   & 44  & 0    \\
    Passages containing no relation        & 35   & -   & -   & -    \\
    Causation                              & 7056 & -   & -   & -    \\
    Conditionality                         & 659  & -   & -   & -    \\
    Hierarchical Relationships             & 1089 & -   & -   & -    \\
    \hline
    \end{tabular}
    \caption{The statistical results of the dataset include the sentence length, the number of relations contained in each sentence, and the specific quantity of each relation.}    \label{tab:statistical}

    \end{table}

\section{Back Translation Implementation}\label{sec:BackTranslation}
Layer 2 provides essential knowledge for classification but may simplify the task, as models may compare two sentences straightforward to judge.
This sublayer transforms representations of knowledge in Layer 2, preserving its inherent meaning.
We exploit \textit{back-translation}~\citep{tan-etal-2019-learning}, additionally necessitating the capability to identify mentions, as original dataset is in Chinese.
Because causal mentions in medical typically follow standard terminologies, receiving fewer modifications during back-translation compared to non-causal contexts.
In practice, we utilize the DeepL API \footnote{\href{https://www.deepl.com/translator}{https://www.deepl.com/translator}} to translate texts retrieved from Langchain (consistent with Layer 2) into English and then directly translate them back.
\section{Discussion of External Knowledge in Layer 3}\label{sec:Effectiveness_Layer3}
\begin{table*}[!ht]
\centering
\small
\begin{tabular}{|p{0.4\linewidth}|p{0.5\linewidth}|}
\hline
\textbf{Sentence} & \textbf{Knowledge} \\
\hline
\begin{CJK*}{UTF8}{gbsn}全身症状表现为精神不振、食欲减退、烦躁不安、轻度腹泻或呕吐\end{CJK*} & \begin{CJK*}{UTF8}{gbsn}小儿时期常见的呕吐是婴幼儿和儿童时期常见的临床症状之一，几乎任何感染或情绪紧张都可引起呕吐\end{CJK*} \\
\textit{General symptoms manifest as malaise, decreased appetite, irritability, mild diarrhea, or vomiting.} & \textit{Vomiting during childhood is a common clinical symptom in infants and children, and can be caused by nearly any infection or emotional stress.} \\
\hline
\begin{CJK*}{UTF8}{gbsn}一旦玻璃体当中水分越来越多,就会造成玻璃体和视网膜发生分离,这就是玻璃体后脱离\end{CJK*} & \begin{CJK*}{UTF8}{gbsn}近视尤其高度近视患者，玻璃体发生液化，纤维化以至后脱离\end{CJK*} \\
\textit{Once the vitreous body accumulates more water, it can lead to a separation between the vitreous body and the retina, known as posterior vitreous detachment.} & \textit{In patients with myopia, especially high myopia, the vitreous body can undergo liquefaction and fibrosis, leading to detachment.} \\
\hline
\begin{CJK*}{UTF8}{gbsn}过度的饮酒,会导致颈部的血管收缩加快,出现其他的一些不必要的并发症\end{CJK*} & \begin{CJK*}{UTF8}{gbsn}长期酗酒每天达100g容易导致向颈段脊髓供血的根动脉缺血\end{CJK*} \\
\textit{Excessive drinking can lead to accelerated constriction of the blood vessels in the neck, resulting in other unnecessary complications.} & \textit{Long-term heavy drinking, reaching 100g per day, can easily cause ischemia in the radicular arteries that supply blood to the cervical spine.} \\
\hline
\end{tabular}
\caption{Examples of Sentences and Corresponding Knowledge}
\label{table:example_sentences_knowledge}
\end{table*}
Table~\ref{table:example_sentences_knowledge} provides examples of the corresponding knowledge provided to the model in Layer 3 when answering questions. To ensure the reliability of the knowledge provided in Layer 3, we randomly selected 50 samples to check whether the additional medical knowledge provided is related to the content of the question or the entities mentioned in the question. In the 50 samples examined, 43 of them had medical entities in the knowledge section that were related to the question statement, while the rest were unrelated. There were 31 samples that provided clear descriptive help for the causal relationship judgment of the question statement, whereas 19 did not offer significant useful information.

\section{Retrieval Augmentation Design}\label{sec:langchain}
\begin{figure*}[!htb]
	\begin{center}
		\includegraphics[width=0.8\linewidth]{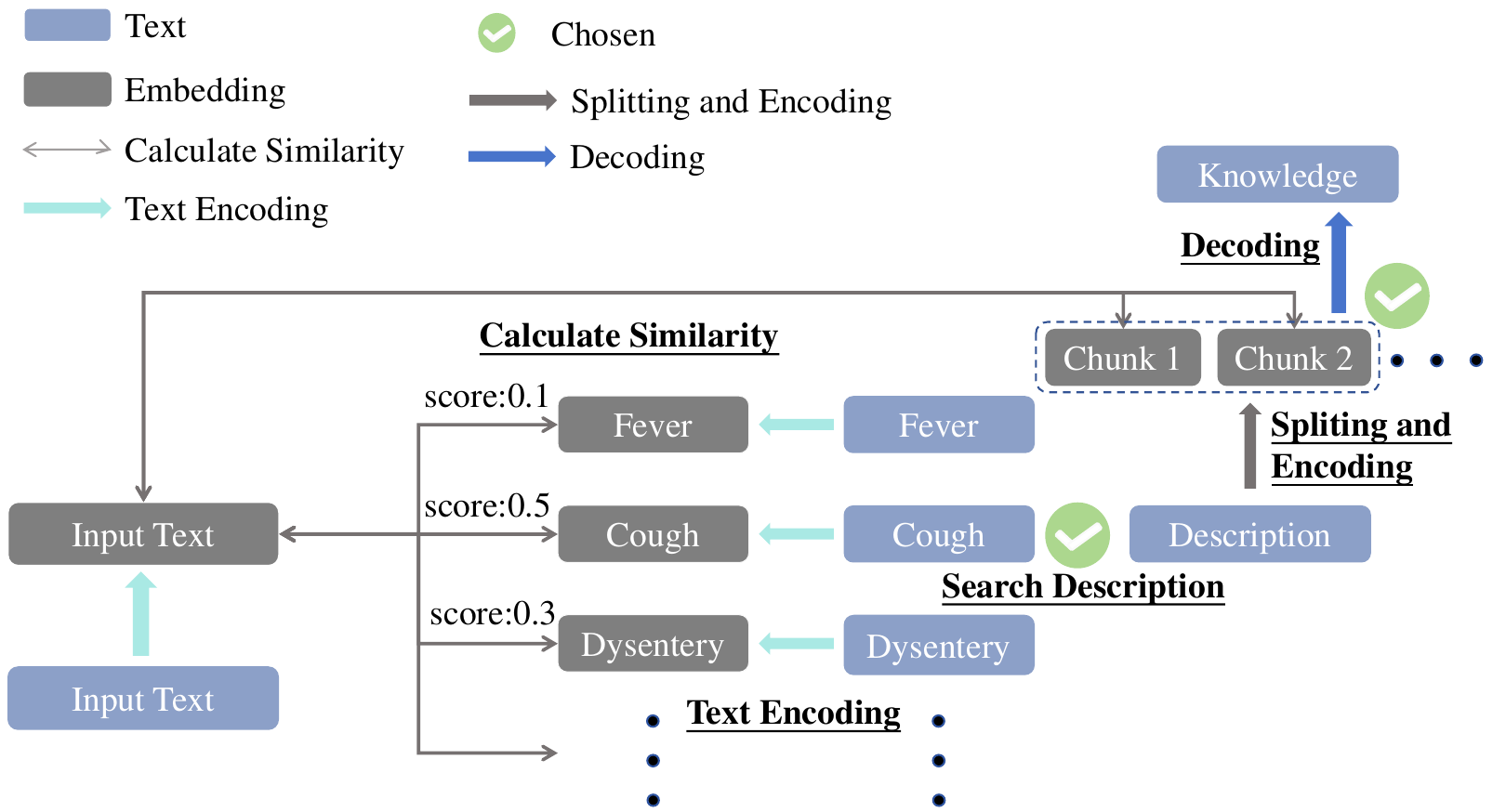}
		\caption{Flowchart of secondary retrieval using langchain, where "encoding" means encoding the input text using the Sentence Transformer, "Calculate Similarity" means calculate the similarity score using the cosine similarity, "Search Description" Indicates the description of the corresponding medical text in the knowledge graph, "Spliting and Encoding" means that the description text is chunked and encoded and "Decoding" means decoding the encoded vector into a sentence.}
		\label{fig:langchain}
	\end{center}
\end{figure*}
To engage retrieval pipeline, we divide each sentence into chunks, allowing for overlap between them, and then encode each chunk using Sentence Transformer \citep{reimers-gurevych-2019-sentence}.
We treat input of LLMs as a query and utilize FAISS (Facebook AI Similarity Search)~\citep{8733051} to efficiently match the encoded query with locally stored sentence vectors, retrieving the top $k$ (set to 2 practically) most relevant chunks.
Since the retrieved sentence fragments may be incomplete, directly providing this knowledge to models would result in receiving inadequate information or incoherent statements.
To address this issue, we control locations of retrieved text fragments in the original text and individually expand the head and tail of the fragment until it forms a complete sentence.
Specifically, due to the large volume of textual data in the medical knowledge graph at Layer 3, directly using Langchain for item-by-item matching is highly inefficient. Therefore, we have adopted a hierarchical retrieval strategy. As shown in Fig.~\ref{fig:langchain}, we first match the input text with disease names in the knowledge graph to select the most relevant diseases. Then, we match the input text with the textual descriptions corresponding to the selected diseases to identify the medical knowledge most relevant to the input text. This selected medical knowledge is ultimately integrated into the external medical knowledge required at Layer 3.

\section{Prompts}\label{sec:Prompt}
\subsection{Structure of Prompts}\label{Structure_of_Prompts}
\begin{table*}[!htb]
\centering
\small
\begin{tabular}
{|p{0.15\linewidth}|p{0.33\linewidth}|p{0.33\linewidth}|}
\hline
Type & Chinese & English \\
\hline
Text & \begin{CJK*}{UTF8}{gbsn}全身症状表现为精神不振、食欲减退、烦躁不安、轻度腹泻或呕吐,全身症状在小宝宝身上可能相对来说比较突出。\end{CJK*} & \textbf{General symptoms include lethargy, decreased appetite, irritability, mild diarrhea, or vomiting. These symptoms may be relatively prominent in babies.} \\
\hline
Retrieved Knowledge & \begin{CJK*}{UTF8}{gbsn}小儿时期常见的呕吐是婴幼儿和儿童时期常见的临床症状之一，几乎任何感染或情绪紧张都可引起呕吐。\end{CJK*} & \textit{Vomiting in childhood is a common clinical symptom in infants and children, which can be caused by almost any infection or emotional stress.} \\
\hline
Simple Prompt & \begin{CJK*}{UTF8}{gbsn}额外医疗为：小儿时期常见的呕吐是婴幼儿和儿童时期常见的临床症状之一，几乎任何感染或情绪紧张都可引起呕吐。根据以上辅助知识和你已知的知识，回答：语句"全身症状表现为精神不振、食欲减退、烦躁不安、轻度腹泻或呕吐,全身症状在小宝宝身上可能相对来说比较突出"因果逻辑正确还是错误。\end{CJK*} & Additional medical knowledge: \textit{Vomiting in childhood is a common clinical symptom in infants and children, which can be caused by almost any infection or emotional stress.} Given the above knowledge and what you know, answer: Is the statement "\textbf{General symptoms include lethargy, decreased appetite, irritability, mild diarrhea, or vomiting, and these symptoms may be relatively prominent in babies}" logically correct or incorrect? \\
\hline
\end{tabular}
\caption{Example of simple prompt}
\label{table:SimplePromptExample}
\end{table*}

\begin{table*}[!htb]
\centering
\small
\begin{tabular}
{|p{0.15\linewidth}|p{0.33\linewidth}|p{0.33\linewidth}|}
\hline
Type & Chinese & English \\
\hline
Text & \begin{CJK*}{UTF8}{gbsn}全身症状表现为精神不振、食欲减退、烦躁不安、轻度腹泻或呕吐,全身症状在小宝宝身上可能相对来说比较突出。\end{CJK*} & \textbf{General symptoms include lethargy, decreased appetite, irritability, mild diarrhea, or vomiting. These symptoms may be relatively prominent in babies.} \\
\hline
Retrieved Knowledge & \begin{CJK*}{UTF8}{gbsn}小儿时期常见的呕吐是婴幼儿和儿童时期常见的临床症状之一，几乎任何感染或情绪紧张都可引起呕吐。\end{CJK*} & \textit{Vomiting in childhood is a common clinical symptom in infants and children, which can be caused by almost any infection or emotional stress.} \\
\hline
Advanced Prompt & [Round 0] \textbackslash n \begin{CJK*}{UTF8}{gbsn}问：你现在在进行句子因果逻辑关系分析的任务\end{CJK*}\textbackslash n\begin{CJK*}{UTF8}{gbsn}答：好的\end{CJK*}\textbackslash n[Round 1]\textbackslash n \begin{CJK*}{UTF8}{gbsn}问：可能会出现因果倒置，涉及到因果关系的对象对应关系错误等错误。\end{CJK*} \textbackslash n \begin{CJK*}{UTF8}{gbsn}答：好的\end{CJK*}\textbackslash n[Round 2]\textbackslash n\begin{CJK*}{UTF8}{gbsn}问：你现在在进行句子因果逻辑关系分析的任务。\end{CJK*}\textbackslash n \begin{CJK*}{UTF8}{gbsn}答：好的\end{CJK*}\textbackslash n[Round 3]\textbackslash n \begin{CJK*}{UTF8}{gbsn}问：这部分是为你提供的额外医疗知识：小儿时期常见的呕吐是婴幼儿和儿童时期常见的临床症状之一，几乎任何感染或情绪紧张都可引起呕吐\end{CJK*}\textbackslash n \begin{CJK*}{UTF8}{gbsn}答：好的\end{CJK*}\textbackslash nquestion: \begin{CJK*}{UTF8}{gbsn}语句："全身症状表现为精神不振、食欲减退、烦躁不安、轻度腹泻或呕吐,全身症状在小宝宝身上可能相对来说比较突出"这个语句是否逻辑正确？先回答是或者否，再给出对应的理由。\end{CJK*} & [Round 0]\textbackslash n Q: You are now performing a task of analyzing the causal logical relationship of sentences.\textbackslash n A: Okay.\textbackslash n[Round 1]\textbackslash n Q: There may be errors such as reversal of cause and effect, involving incorrect object correspondence of causal relationships.\textbackslash n A: Okay.\textbackslash n [Round 2]\textbackslash n Q: You are now performing a task of analyzing the causal logical relationship of sentences.\textbackslash n A: Okay.\textbackslash n [Round 3]\textbackslash n Q: This part is to provide you with additional medical knowledge: \textit{Vomiting in childhood is a common clinical symptom in infants and children, which can be caused by almost any infection or emotional stress.}\textbackslash n A: Okay.\textbackslash n Question: Is the statement "\textbf{General symptoms include lethargy, decreased appetite, irritability, mild diarrhea, or vomiting, and these symptoms may be relatively prominent in babies}" logically correct? Answer yes or no, then provide the corresponding reason. \\
\hline
\end{tabular}
\caption{Example of advanced prompt}
\label{table:AdvancedPromptExample}
\end{table*}
Prompts for probing were designed according to \textit{Base Prompt Framework} by Elvis Saravia\footnote{\href{https://www.promptingguide.ai/}{https://www.promptingguide.ai/}}.
All prompts have the following elements: \textbf{Instructions}, we instruct LLMs with a binary classification tasks.
\textbf{Contexts}, we place supplementary knowledge and contexts in this section for problems contexts. In the layer of bare asking, this part is excluded.
\textbf{Input Data}, we place sentence to be classified in this slot, separated with Chinese quotation mark.
\textbf{Output Indicator}, we instruct models about output format and order, the best indicator is to make classification first and then explain why.
Extensive search for other prompts is neglected, since we consider understanding of reasonable prompts to be part of models capabilities.

\subsection{Examples of Prompts}\label{Example_of_Prompts}
When using a simple prompt, we directly connect the additional knowledge with the question content in a straightforward manner, as illustrated by the example in Table~\ref{table:SimplePromptExample}. In contrast, when using an advanced prompt, we employ multi-turn dialogues to emphasize the task content and separate the parts that provide knowledge from those that pose questions. This approach allows the model to understand the task content, and the boundaries between knowledge and questions more clearly. Examples of this can be found in Table~\ref{table:AdvancedPromptExample}.

\section{Details of Models}\label{sec:Models}
\paragraph{GPT-4}~\citep{DBLP:journals/corr/abs-2303-08774}.
We use a static version of \textit{GPT-4-0613}~\footnote{\href{https://platform.openai.com/docs/models/gpt-4}{https://platform.openai.com/docs/models/gpt-4}} for experiment.

\paragraph{GPT-3.5}~\citep{NEURIPS2022_b1efde53}. We use \textit{GPT-3.5-Turbo}\footnote{\href{https://platform.openai.com/docs/models/gpt-3-5}{https://platform.openai.com/docs/models/gpt-3-5}} static version of ChatGPT.

\paragraph{ChatGLM}~\citep{DBLP:conf/iclr/ZengLDWL0YXZXTM23,du-etal-2022-glm}. It is pretrained mainly on Chinese and English corpus, and can recognize Chinese expressions better.

\paragraph{MedChatGLM} is a model under fine-tuning on ChatGLM in Chinese medical corpus.

\paragraph{BERT}~\citep{devlin-etal-2019-bert}~\footnote{\href{https://huggingface.co/bert-base-chinese}{https://huggingface.co/bert-base-chinese}} is trained on supervised datasets, classification is extracted using masked language model (MLM), in which BERT is trained to fill certain slot with \textit{right} or \textit{wrong}.
\section{Performance Difference of LLMs}\label{ModelSelection}
The performance of action 3 is the worst of the three and has even approached random categorization for some models (ChatGLM~\citep{DBLP:conf/iclr/ZengLDWL0YXZXTM23,du-etal-2022-glm} and MedChatGLM).
The assistance of original passage causes a performance drop, regardless prompt used for all models including GPT-4~\citep{DBLP:journals/corr/abs-2303-08774}.
In contrast, the introduction of layer 2 in action 1 improves performance.
This means that model lacks understanding of causation between mentions, relying more on linguistic order and positions. We believe that the ability to judge causal relevance problems is mainly related to the number of model parameters and the training method.

\paragraph{Training Strategies}
 GPT-4 and GPT-3.5 uses the RLHF~\citep{NEURIPS2022_b1efde53} training strategy, which makes its answer results more similar to human beings. This can improve the logic of its dialogue and improve its ability to discuss causal problems to a certain extent. 

 \paragraph{The Number of Model Parameters}
 Compared with GPT-3.5 and ChatGLM, GPT-4 has a larger number of parameters and a larger knowledge reserve, and it has a stronger ability to understand complex logic.
\section{PPL of Positive and Negative Instances}\label{sec:ppl}
This section presents the specific experimental results of testing various actions using GPT-2 Chinese\footnote{\href{https://huggingface.co/uer/gpt2-chinese-cluecorpussmall}{https://huggingface.co/uer/gpt2-chinese-cluecorpussmall}}~\citep{radford2019language} to determine the confidence of sentences based on PPL.
We test PPL on all actions of datasets, and compare difference of positive and negative instances.
When difference is big, dataset are more easier for classification from statistical association.

\begin{figure}[!ht]
	\begin{center}
		\includegraphics[scale=0.35]{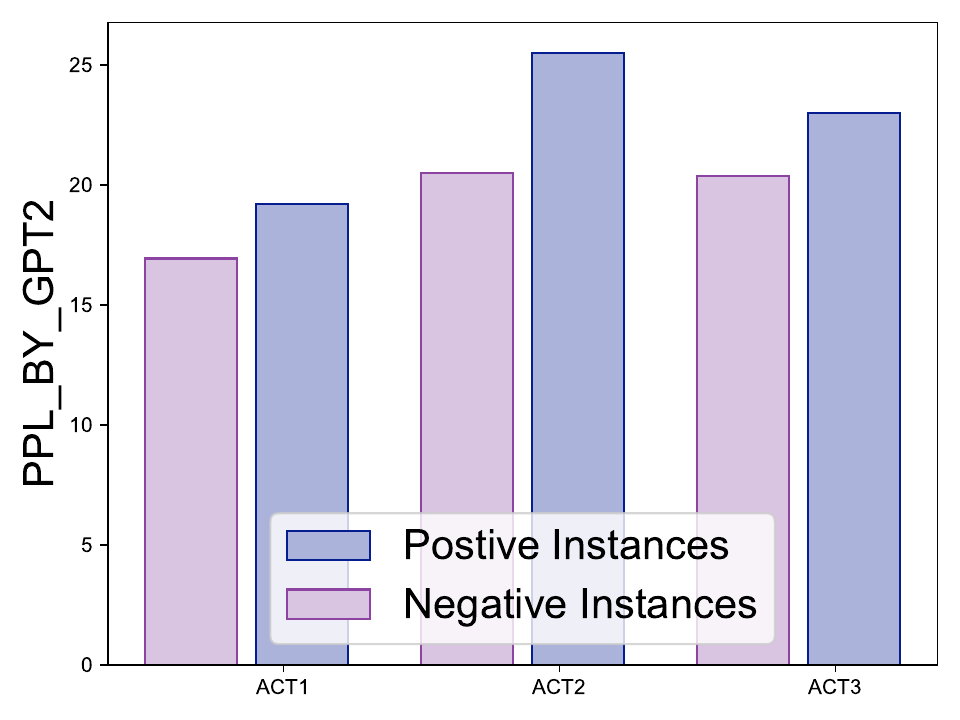}
		\caption{PPL of positive and negative instances in different actions calculated by GPT-2}
		\label{fig:ppl}
	\end{center}
\end{figure}

As shown in Fig.~\ref{fig:ppl}.
PPL is correlated with the model's confidence in a given sentence using statistical associations.
Results show that action 2 is more easily distinguishable statistically, with a higher base PPL and a more pronounced increase in negative instances.
\end{document}